\documentclass[11pt,a4paper]{article}
\usepackage[hyperref]{acl2018}
\usepackage{times}
\usepackage{latexsym}

\usepackage{amsmath,amssymb}
\usepackage{graphicx}
\usepackage{multirow}
\usepackage[utf8]{inputenc}

\usepackage{float}
\usepackage{footnote}
\usepackage{url}

\usepackage[normalem]{ulem}
\definecolor{darkgreen}{rgb}{0.0, 0.5, 0.0}

\def\ODdel#1{\bgroup\markoverwith{\textcolor{darkgreen}{\rule[0.5ex]{2pt}{1pt}}}\ULon{#1}}

\def\YKdel#1{\bgroup\markoverwith{\textcolor{blue}{\rule[0.5ex]{2pt}{1pt}}}\ULon{#1}}

\def\SAdel#1{\bgroup\markoverwith{\textcolor{red}{\rule[0.5ex]{2pt}{1pt}}}\ULon{#1}}

\def\VRdel#1{\bgroup\markoverwith{\textcolor{magenta}{\rule[0.5ex]{2pt}{1pt}}}\ULon{#1}}

\usepackage{array}
\usepackage{arydshln}
\newcolumntype{L}[1]{>{\raggedright\arraybackslash}m{#1}}
\newcolumntype{C}[1]{>{\centering\arraybackslash}m{#1}}
\newcolumntype{R}[1]{>{\raggedleft\arraybackslash}m{#1}}

\aclfinalcopy 

\title{A Knowledge-Grounded Multimodal Search-Based Conversational Agent}

\author{Shubham Agarwal *,\ \ Ondřej Dušek, Ioannis Konstas and Verena Rieser \\ 
The Interaction Lab,
Department of Computer Science\\
Heriot-Watt University, 
Edinburgh, UK\\
* Adeptmind Scholar, Adeptmind Inc., Toronto, Canada\\
\texttt{\{sa201, o.dusek, i.konstas, v.t.rieser\}@hw.ac.uk} 
}

\date{}

\begin{document}
\maketitle
\begin{abstract} 
Multimodal search-based dialogue is a challenging new task: It extends visually grounded question answering systems into multi-turn conversations with access to an external database. We address this new challenge by learning a neural response generation system from the recently released Multimodal Dialogue (MMD) dataset \cite{saha2017multimodal}. We introduce a knowledge-grounded multimodal conversational model where an encoded knowledge base (KB) representation is appended to the decoder input. Our model substantially outperforms strong baselines in terms of text-based similarity measures (over 9 BLEU points, 3 of which are solely due to the use of additional information from the KB).
\end{abstract}

\section{Introduction}
\label{introduction}

Conversational agents have become ubiquitous, with variants ranging from open-domain conversational chit-chat bots  \cite{ram2018conversational,papaioannou2017alana,fang2017sounding} to domain-specific task-based dialogue systems \cite{singh2000reinforcement,rieser2010natural,rieser2011reinforcement,young2013pomdp,wen2016network}.

\begin{figure}[ht]
{\centering
\includegraphics[scale=0.5]{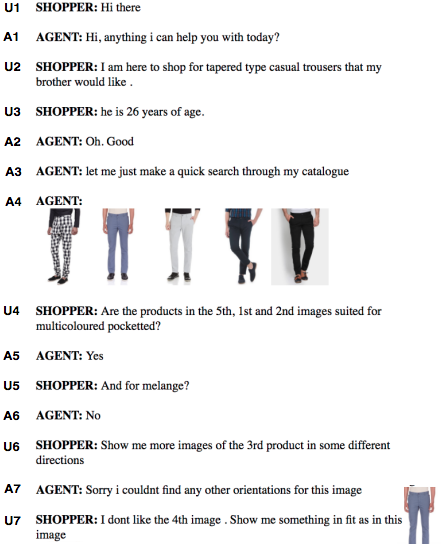} 
}
\vspace{-7mm}
\caption{Example chatlog depicting multimodal user-agent interaction in a dialogue session from the MMD dataset. 
The system needs to ground knowledge to generate responses related to product-specific attributes. We focus on textual response generation given a fixed-size conversational history.}
\label{fig:task}
\vspace{-5mm}
\end{figure}

Our work builds upon the recently released Multimodal Dialogue (MMD) dataset \cite{saha2017multimodal}, which contains dialogue sessions in the e-commerce (fashion) domain.  Figure \ref{fig:task} illustrates an example chat session with multimodal interaction between the user and the system. We focus on the task of generating textual responses  conditioned on the previous conversational history. Traditional goal-oriented dialogue systems relied on slot-filling approach to this task, i.e.\ explicit modelling of all attributes in the domain \cite{lemon2006isu,wang2013simple,young2013pomdp}. On the other hand, previous work on MMD data used direct learning from raw texts with implicit semantic representation only. This paper attempts to combine both approaches by learning to generate replies from raw user input, while also incorporating Knowledge Base (KB) inputs (i.e.\ explicit semantics) into the generation process. We discuss how our model is able to handle various user intents (request types) and the impact of incorporating the additional explicit semantic information from the KB into particular targeted intents. We use user intent annotation and KB queries provided with the dataset for the purpose of this work.

Our main contribution is the resulting fully data-driven model for the task of conversational multimodal dialogue generation, grounded in conversational text history, vision and KB inputs. We also illustrate a method to improve context modelling over multiple images and show great improvements over the baseline. Finally, we present a detailed analysis of the outputs generated by our system corresponding to different user intents.

\section{Related Work}

With recent progress in deep learning, there is continued interest in the tasks involving both vision and language, such as image captioning \cite{xu2015show,vinyals2015show,karpathy2015deep}, visual storytelling \cite{ferraro2016visual}, video description \cite{venugopalan2014translating,venugopalan2015sequence} or dialogue grounded in visual context \cite{antol2015vqa,das2017visual,tapaswi2016movieqa}.

\newcite{bordes2016learning} and \newcite{ghazvininejad2017knowledge} presented knowledge-grounded neural models; however, these are uni-modal in nature, involve only textual interaction and do not take into account the conversational history in a dialogue. In contrast, our system grounds on a KB while also conditioning on previous dialogue context which is multimodal in nature, consisting of both textual and visual communication between the user and the system. We formulate our KB input from a database query (triggered by the system) similar to \newcite{sha2017order}, as described in Section~\ref{sect:KB}.

Our model belongs to the encoder-decoder paradigm 
where sequence-to-sequence models \cite{cho2014learning,sutskever2014sequence,bahdanau2014neural} have become the de-facto standard for natural language generation. However, they tend to ignore the conversational history in a dialogue. 
The Hierarchical Recurrent Encoder Decoder (HRED) architecture \cite{serban2016building,serban2017hierarchical,lu2016hierarchical} addresses this limitation by using a context recurrent neural network (RNN), forming a hierarchical encoder. We build upon these HRED models and refer to them as Text-only HREDs (T-HRED) in the following. 
Our model is most similar to the Multimodal HRED (M-HRED) of \newcite{saha2017multimodal}, with context and KB extensions (see Section~\ref{sect:model}).

\section{Knowledge grounded Multimodal Conversational model}
\label{sect:model}
\begin{figure}[t]
 \centering
\includegraphics[scale=0.35]{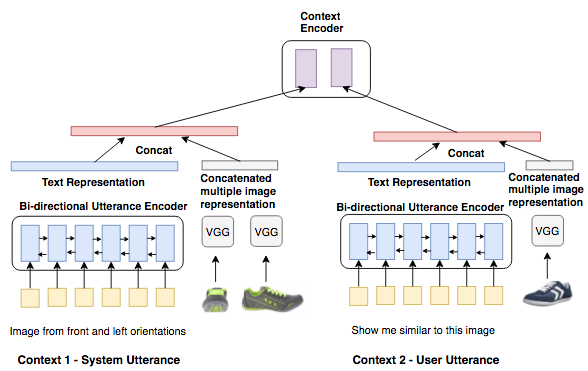}
\vspace{-3mm}
\caption{Schematic diagram of hierarchical encoder described in Section \ref{sect:encoders}. Figure \ref{fig:kb_multimodal} depicts  full pipeline of the model using knowledge base input. In contrast to \citet{saha2017multimodal}, we model over multiple images in a contextual dialogue turn by combining all `local' representations of multiple images to a `global' image representation per turn. We show a context of 2 turns for simplicity.}
\label{fig:model}
\end{figure}

While \newcite{saha2017multimodal} propose Multimodal HRED (M-HRED) by extending T-HRED to include visual context over images, they do not ground their dialogue context over an external database. Also, they limit the visual information by `unrolling' multiple images to just use the last image of a single turn. For example in Figure~\ref{fig:task}, \newcite{saha2017multimodal} consider only the last image of trousers as visual context in Agent's 
response A4. In contrast, we include all the images in a single turn using a linear layer (see \newcite{agarwalimproving2018} for a detailed analysis). 

\begin{figure*}[ht!]
\centering
\includegraphics[scale=0.48]{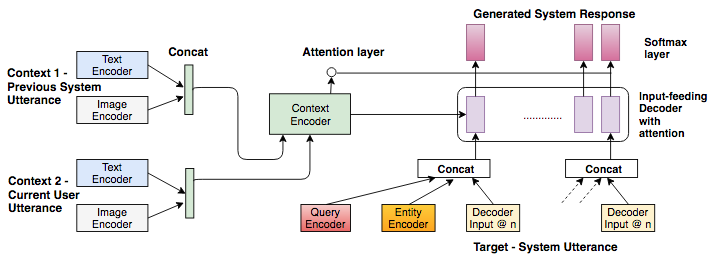}
\vspace{-5mm}
\caption{The full encoder-decoder pipeline of our model. 
While we have early fusion of textual and image representations (which act as input to the context encoder, see Figure~\ref{fig:model}), we employ late fusion of the knowledge base vector at the decoder level. For simplicity, we show a context of 2  turns.}
\label{fig:kb_multimodal}
\vspace{-5mm}
\end{figure*}

In addition, we devise a mechanism to ground our textual responses on a KB; Figure~\ref{fig:kb_multimodal} depicts the full pipeline of our model. We combine textual and visual representations at the encoder level and pass it through the HRED's context encoder (cf.\ Figure~\ref{fig:model}), which learns the backbone of the conversation (see Section~\ref{sect:encoders}). Subsequently, we inject knowledge from the KB at the decoder level in each timestep (see Sections~\ref{sect:KB} and~\ref{decoder}).

Formally, we model a dialogue as a sequence of utterances (turns) which are considered as sequences of words and images: 
\vspace{-2mm}
\begin{equation}
\label{eq:1}
P_{\theta} (t_{1}, \dots t_{N})  = \prod_{n=1}^{N} P_{\theta} (t_{n} | t_{<n} ) 
\end{equation}
Here $t_n$ represents the $n$-th utterance in a dialogue. The whole model is trained using cross entropy on next-word prediction: 
\vspace{-2mm}
\begin{equation}
J(\theta)=-\sum_{n=1}^{N} log P(y_n|y_0 \dots y_{t-1})
\label{eq:loss}
\end{equation}
In the following, we explain all the different components of our model. We use the following notation: $f_{\theta}^{text}$,$f_{\theta}^{cxt}$, $f_{\theta}^{query}$, $f_{\theta}^{ent}$ and $f_{\theta}^{dec}$ are all GRU cells \cite{cho2014learning} and $g_{\theta}^{enc}$ is a Convolutional Neural Network (CNN) image encoder. $\theta$ represent our model weights. $w_{n,m}$ is the $m$-th word in the $n$-th textual utterance. Similarly, $q_{m,n}$ and $c_{m,n}$ represent input at each timestep in the query and entity encoder (see Section~\ref{sect:KB}).

\vspace{-0.5mm}
\subsection{Hierarchical Encoder}
The encoder is formed of the following modules: 
\label{sect:encoders}
\paragraph{Utterance (Text) encoder:}
We pass each utterance (previous system responses as well as current user query)
in a given context through a text encoder. We use bidirectional GRU ($f_{\theta}^{text}$) to generate the textual representation $h_{n,M_{n}}^{text}$ (cf.~Eq.~\eqref{eqn:textEnc}). These textual representations are combined with image representations in each turn, forming the input for the context encoder.
\begin{align}
h_{n,m}^{text} &= f_{\theta}^{text}(h_{n,m-1}^{text}, w_{m,n}) ;\quad
h_{n,0}^{text} = 0 
\label{eqn:textEnc}
\end{align}
\paragraph{Image encoder:}
\label{imageEnc}
We first extract the `local' image representations for all images in a dialogue turn (denoted by $g_{\theta}^{enc}(img_{k})$ in Eq.~\eqref{eqn:imgEnc}) and concatenate them together.\footnote{We used the VGGnet \cite{simonyan2014very} CNN to obtain the local image representations. Since the number of images in a turn is $\leq 5$, we consider  zero vectors in the absence of images.} This concatenated vector is passed through a linear layer to form the `global' image context for a single turn, denoted by $h_{n}^{img}$. 
\begin{align}
h_{n}^{img} &= l^{img} ([g_{\theta}^{enc}(img_{1}), \dots g_{\theta}^{enc}(img_{k})]) \label{eqn:imgEnc} 
\end{align}
\paragraph{Context encoder:}
\label{sect:contextEnc}
The final hidden representations from both text encoder $h_{n,M_{n}}^{text}$ and image encoder $h_{n}^{img}$ are concatenated for each turn and serve as input to the context RNN (cf.~Eq.~\eqref{eqn:cxtEnc}). 
On top of the
text and image encoder, this builds a hierarchical encoder modelling the dialogue history. The final hidden state of the context RNN $h_{N}^{cxt}$ acts as the initial state of the decoder RNN defined in Section~\ref{decoder}. 
\begin{align}
h_{n}^{cxt} &= f_{\theta}^{cxt}(h_{n-1}^{cxt}, [h_{n,M_{n}}^{text}, h_{n}^{img}]);\ h_{0}^{cxt}=0  
\label{eqn:cxtEnc}
\end{align}

\subsection{Knowledge base (KB) input}
\label{sect:KB}

The KB vector $h_{n}^{kb}$ in Eq.~\eqref{eqn:kbVec} is formed by concatenating the $h_{n}^{query}$ and $h_{n}^{ent}$ representations. While our approach is modelled around the MMD dataset which provides contextual KB queries and profiles of celebrities endorsing specific products, it can be applied to other KBs with encoded queries and (optionally)  properties of relevant entities.
\vspace{-1mm}
\begin{align}
h_{n}^{query} &= f_{\theta}^{query}(h_{n-1}^{query}, q_{m,n}) \label{eqn:queryEnc}\\
h_{n}^{ent} &= f_{\theta}^{ent}(h_{n-1}^{ent}, c_{m,n}) \label{eqn:celebEnc}\\
h_{n}^{kb} &= [h_{n}^{query},h_{n}^{ent}] \label{eqn:kbVec}\\
h_{0}^{query} &= 0 ;\quad h_{0}^{ent} = 0
\end{align}
\paragraph{Query encoder:}
\label{kbEncoder}
Each chat session contains multiple queries to the database which retrieve the relevant product suited to user requirements at specific turn. We replicate this query for subsequent dialogue turns until a new query is triggered by the system. This query acts as knowledge base for the model at each turn. We show a sample input to the model in Figure \ref{kbInput}. We used unidirectional GRU cell to encode the query input $h_{n}^{query}$. 

\begin{figure}[tb]
\footnotesize
\textbf{Query:}\vspace{-2mm}
\begin{verbatim}
"search_criteria": {
    "name": {"driving shoes": 1.0}, 
    "fit":{"tight": 1.0}, 
    "brand": {"cirohuner": 1.0},
    "image_type":{"front": 1.0}, 
    "gender": {"men": 1.0}, 
    "print": {"chain": 1.0}
} 
\end{verbatim}
\textbf{Knowledge base input:}\vspace{-2mm}
\begin{verbatim}
name driving shoes fit tight brand
cirohuner image_type front gender
men print chain
\end{verbatim}
\vspace{-3mm}
\caption{Sample query to the database and corresponding knowledge base input vector.}
\label{kbInput}
\end{figure}
\vspace{-3mm}
\paragraph{Entity encoder:}
\label{sect:celebEnc}
The input to the entity encoder is a list of entities relevant to the query at hand (see Figure~\ref{celebInput}). GRU cells are used to produce the resulting $h_n^{ent}$. Specifically, the MMD dataset categorises products into synonym sets (synsets) and provides a list of celebrities endorsing each synset (see Section~\ref{sect:dataset} for details).

This input is used specifically for the `celebrity' intent in our model, where the user asks about celebrities endorsing a product.
For each target prediction with celebrity intent, we first extract the relevant celebrity profiles  using basic pattern matching over the user utterance. For each of the celebrities in the user query, we order the corresponding synsets by their probability of endorsement. If no celebrity is found, we use synset information from the query to extract celebrities which endorse the corresponding synset.

\begin{figure}[tb]
\footnotesize
\textbf{1.}\vspace{-2mm}
\begin{verbatim}
User: what kind of trousers are 
endorsed by celebrity cel_237?
Intent: celebrity
Subintent :does_celebrity_endorse_n 
Celebrity: cel_237
Celebrity input: boxer briefs 
\end{verbatim}
\textbf{2.}\vspace{-2mm}
\begin{verbatim}
User: which of the celebrities 
usually wear similar looking canvas 
shoes as in the 2nd image
Intent: celebrity
Subintent: which_celebrity_endorses_n
Synset: canvas shoes
Celebrity input: 
cel_987 cel_2 cel_316 cel_101 
\end{verbatim}
\vspace{-3mm}
\caption{Two input scenarios for the entity encoder depending on the fine grained user intent. If there is no `celebrity' intent, we have an empty string as input to the entity encoder.}
\label{celebInput}
\end{figure}
\subsection{Input feeding decoder}
\label{decoder}
We use an input feeding decoder with the attention mechanism of \newcite{luong2015effective}. 
We concatenate the KB input $h_n^{kb}$ with the decoder input (cf.\ Eq.~\eqref{eqn:dec}, where $h_{n,0}^{dec}=h_{N}^{cxt}$). The rationale behind this late fusion of KB representation is that KB input remains the same for a given context and does not change on each turn. On the other hand, images and textual response together form a context in a dialogue turn and thus we fuse them early at the encoder level. The decoder is trained using cross-entropy loss defined in Eq.~\eqref{eq:loss}.
\begin{align}
h_{n,m}^{dec}&=f_{\theta}^{dec}(h_{n,m-1}^{dec}, w_{n,m} ,h_{n-1}^{cxt},h_{n-1}^{kb}) \label{eqn:dec}
\end{align}

\section{Experiments and Results}
\label{sect:DatasetImplementationMetrics}
\subsection{Dataset}
\label{sect:dataset}
Our work is based on the Multimodal Dialogue (MMD) dataset \cite{saha2017multimodal}, which consists of 150k chat sessions.\footnote{We used the same training-development-test split as provided by the dataset authors.} 
User queries can be complex from the perspective of multimodal task-specific dialogue, such as ``Show me more images of the 3rd product in some different directions''. However, it also heavily relies on the external KB to answer product attributes related to user queries, such as ``What is the brand/material of the suit in 3rd image?'' or ``Show something similar to 1st result but in a different material''. This dataset contains raw chat logs as well as metadata information of the corresponding products. Around 400 anonymised celebrity profiles have been introduced in the system to emulate endorsement in recommendation, such as ``What kind of slippers are endorsed by cel\_145?''. For each dialogue turn, there are manual annotations of the user intent available. We use the intents to construct celebrity encodings. On average, each session contains 40 dialogue turns. The system response depends on the intent state of the user query and on average contains 8 words and 4 images per utterance. We created our own version of the dataset from the raw chat logs of the dialogue session and metadata information. As discussed in Section~\ref{sect:encoders}, this was necessary to model the visual context over multiple images. We created the KB input to our model as described in Section~\ref{kbEncoder} from the raw chat logs and the metadata information. 

\subsection{Implementation}
We used PyTorch\footnote{\url{https://pytorch.org/}} \cite{paszke2017automatic} for our experiments.\footnote{Code can be found at:\\ \url{https://github.com/shubhamagarwal92/mmd}}
We did not use any kind of delexicalisation\footnote{Replacing specific values with placeholders \cite{henderson_robust_2014}.} and rely on our model to directly learn from the conversational history and KB. All encoders and decoders are based on 1-layer GRU cells \cite{cho2014learning} with 512 as the hidden state size. We used the 4096 dimensional FC6 layer image representations from VGG-19 \cite{simonyan2014very} provided by \newcite{saha2017multimodal}. Adam \cite{kingma2014adam} was chosen as the optimizer, and we clipped gradients greater than 5. We experimented with different learning rates and settled on the value of 0.0004. Dropout of 0.3 is applied to all the RNN cells to avoid overfitting, and we perform early stopping by tracking the validation loss (with single trial for each experiment).

\subsection{Analysis and Results}
\label{sect:analysis}

We evaluate our response generation using the \textsc{Bleu} \cite{papineni2002bleu}, \textsc{Meteor} \cite{lavie2007meteor} and \textsc{Rouge-L} \cite{lin2004automatic} automatic metrics.\footnote{We used the evaluation scripts provided by \cite{sharma2017nlgeval}.} We reproduce the baseline results from \newcite{saha2017multimodal} using their code and data-generation scripts.\footnote{\url{https://github.com/amritasaha1812/MMD_Code}}

\begin{table}[ht]
{\centering
\resizebox{0.49\textwidth}{!}{
\begin{tabular}{l|c|cccccc}
  \hline
Model & Cxt & \textsc{Bleu-4} &\textsc{Meteor} & \textsc{Rouge-L} \\
  \hline \hline
Saha et al. M-HRED* & 2 & 0.3767 & 0.2847 & 0.6235  \\\hdashline[0.5pt/2pt]
T-HRED & 2 & 0.4292 & 0.3269 & 0.6692\\
M-HRED & 2 & 0.4308 & 0.3288 & 0.6700\\
T-HRED--attn & 2 & 0.4331 & 0.3298 & 0.6710\\
M-HRED--attn & 2 &  0.4345 & 0.3315 & 0.6712\\
T-HRED--attn & 5 & 0.4442 & 0.3374 & 0.6797\\
M-HRED--attn & 5  & 0.4451 & 0.3371 & 0.6799 \\\hdashline[0.5pt/2pt]
M-HRED--kb & 2 & 0.4573 & 0.3436 & 0.6872 \\
T-HRED--attn--kb & 2 & 0.4601 & 0.3456 & 0.6909 \\
M-HRED--attn--kb & 2 & 0.4624 & 0.3476 & 0.6917 \\
T-HRED--attn--kb & 5 & 0.4612 & 0.3461 & 0.6913 \\
M-HRED--attn--kb & 5 & \textbf{0.4634} & \textbf{0.3480} & \textbf{0.6923} \\
\hline
\end{tabular}
}}
\vspace{-2mm}
\caption{Automated evaluation based on \textsc{BLEU-4}, \textsc{METEOR} and \textsc{ROUGE-L} metrics. Here, `M' represents multimodality while `T' stands for text-only model. `attn' denotes use of attention and `kb' signifies incorporating Knowledge Base input. `Cxt' represents context size for the dialogue history.\\ 
*Saha et al.\ was trained on a different version of the dataset, as discussed in Section \ref{sect:model}.}
\label{table:our_results}
\end{table}

\begin{table}[ht!]
\centering
\small
\begin{tabular}{l|ll}
  \hline
Intent & Model & \textsc{Bleu-4} \\
  \hline \hline
\multirow{2}{*}{show-similar-to}& 
M-HRED--attn & 0.9998 \\
& M-HRED--attn--kb & 1.0 \\ \hline
\multirow{2}{*}{sort-results}& 
 M-HRED--attn &  0.9188 \\
& M-HRED--attn--kb & 0.9384 \\ \hline
\multirow{2}{*}{suited-for}&  M-HRED--attn & 0.6151  \\ 
& M-HRED--attn--kb & 0.6216 \\ \hline
\multirow{2}{*}{show-orientation}&  M-HRED--attn & 0.5388 \\ 
& M-HRED--attn--kb & 0.5854 \\ \hline
\multirow{2}{*}{buy}&  M-HRED--attn & 0.2665 \\ 
& M-HRED--attn--kb & 0.3179 \\ \hline
\multirow{2}{*}{ask-attribute}&  M-HRED--attn & 0.4960 \\ 
& M-HRED--attn--kb & 0.5934  \\ \hline
\multirow{2}{*}{celebrity}
& M-HRED--attn & 0.2671 \\ 
& M-HRED--attn--kb & 0.2725 \\ \hline
\end{tabular}
\vspace{-2mm}
\caption{
BLEU scores for the entire corpus predictions for specific intents with a context of 5.
}
\label{table:intentBleu}
\vspace{-5mm}
\end{table} 

Table \ref{table:our_results} summarises the results for our M-HRED model without incorporating KB information. Attention-based models consistently outperform their counterparts. Adding the visual inputs does not lead to major improvements (M-HRED vs.\ T-HRED for a given context).
However, grounding in KB gave a stark uplift (M-HRED--attn--kb vs.\ M-HRED--attn) for a given context size. Adding KB input boosts performance more for a shorter context compared to longer context. It can be conjectured that the longer context contains some of the information that is in the KB queries and so there is less impact of the KB input when we include the longer context. Compare the difference for M-HRED--attn--kb vs.\ M-HRED--attn for a context of 2 (3 BLEU points) vs. 5 (2 BLEU points) in Table \ref{table:our_results}. Conversely, longer context improves more the models without KB queries. 

In summary, our best performing model (M-HRED--attn--kb) outperforms the model of \citet{saha2017multimodal} by 9 \textsc{Bleu} points. We also analysed our generated outputs for different user intents, as shown in Table~\ref{table:intentBleu}. As assumed, intents such as `show-similar-to' and `sort-results' are relatively easy from the perspective of NLG, requiring no information about the product description; our model matches the reference almost perfectly. 

\begin{table*}[ht!]
\centering
\resizebox{\textwidth}{!}{
\begin{tabular}{C{2.5cm}|L{3cm}L{14cm}}
  \hline
Intent & Model & Example Text\\
  \hline \hline
\multirow{5}{*}{show-similar-to}&  & \textbf{Text context}: yes. $|$ show me something similar to the 1st image but in a different material \\
 & & \textbf{Gold Target:} the similar looking ones are \\ 
  & & \textbf{KB:} \textit{name[casual-trousers] gender[women] brand[antigravity] synsets[casual-trousers]}  \\ 
 \cline{2-3}
& M-HRED--attn & \textbf{Predicted:} the similar looking ones are \\ \cline{2-3}
& M-HRED--attn--kb & \textbf{Predicted:} the similar looking ones are \\ \hline
\multirow{5}{*}{sort-results}& &  \textbf{Text context}: sorry i dont seem to have anything in loop but would you like in slim $|$ sort these by best seller ranking \\
 & & \textbf{Gold Target:} sorry i do not have the best seller ranking information for these \\ 
  & & \textbf{KB:} \textit{name[casual-trousers] gender[women] brand[antigravity] synsets[casual-trousers]}  \\ 
    \cline{2-3}
& M-HRED--attn  & \textbf{Predicted:} sorry i dont have the best seller ranking information for the rest \\ \cline{2-3}
& M-HRED--attn--kb  & \textbf{Predicted:} sorry i do not have the best seller ranking information for these \\ \hline
\multirow{5}{*}{suited-for}& &   \textbf{Text context}: no. $|$ which care might go well with this dupatta? \\
& &  \textbf{Gold Target:} it will suit well for professional and handwash care \\ 
   & & \textbf{KB:} \textit{name[dupatta] gender[women] synsets[dupatta] image-type[front] care[handwash]} \\ 
    \cline{2-3}
& M-HRED--attn  & \textbf{Predicted:} it will suit well for gentle hand care \\ \cline{2-3}
& M-HRED--attn--kb  & \textbf{Predicted:} it will suit well for \textit{hand wash care} \\ \hline
\multirow{5}{*}{show-orientation}& &   \textbf{Text context}: $|$ show me more images of the 3rd product from some different orientations \\
& &  \textbf{Gold Target:} sorry i couldnt find any other orientations for this image \\ 
  & & \textbf{KB:} \textit{name[casual-trousers] gender[women] brand[antigravity] synsets[casual-trousers]}  \\ 
    \cline{2-3}
& M-HRED--attn  & \textbf{Predicted:} image from the front, right, back and left orientations respectively \\ \cline{2-3}
& M-HRED--attn--kb  & \textbf{Predicted:} sorry i couldnt find any other orientations for this image \\ \hline
\multirow{5}{*}{buy}& &   \textbf{Text context}: $|$ i like the 4th one. i think ill buy that one \\
& &  \textbf{Gold Target:} absolutely , thats a great kurta \\ 
  & & \textbf{KB:} \textit{name [kurta] color [green] gender [men] synsets [kurta] image-type [front]} \\ 
\cline{2-3}
& M-HRED--attn  & \textbf{Predicted:} absolutely , i think thats a great jeans \\ \cline{2-3}
& M-HRED--attn--kb  & \textbf{Predicted:} absolutely , i think thats a great \textit{kurta} \\ \hline
\multirow{5}{*}{ask-attribute}&   & \textbf{Text context}: yes. $|$ what is the brand in the 1st result? \\
& &  \textbf{Gold Target:} the blouse in the 1st image has alfani brand \\ 
  & & \textbf{KB:} \textit{name [blouse] brand [alfani] synsets [blouse] image-type [look] gender [women]}
 \\ 
\cline{2-3}
& M-HRED--attn  & \textbf{Predicted:} the brand in 1st image is topshop \\ \cline{2-3}
& M-HRED--attn--kb  & \textbf{Predicted:} the brand in 1st image is \textit{alfani} \\ \hline
\multirow{5}{*}{celebrity}& &   \textbf{Text context}: yes. celebrities cel\_779, cel\_10 and cel\_513 also endorse this type of cufflinks $|$ and celebrity cel\_603 for the 1st? \\
& &  \textbf{Gold Target:} yes \\    & & \textbf{KB Query:} \textit{name[casual-trousers] gender[women] synsets[casual-trousers]} \\ 
 & & \textbf{KB Entity:} \textit{scarf earrings casual trousers casual shirt} \\ 
    \cline{2-3}
& M-HRED--attn  & \textbf{Predicted:} no. \\ \cline{2-3}
& M-HRED--attn--kb  & \textbf{Predicted:} yes. \\ \hline
\end{tabular}
}
\vspace{-4.5mm}
\caption{
Examples of predictions corresponding to different user intents, showcasing the effect of grounding in KB. We show textual context as well as relevant knowledge base input (and omit image context) for brevity's sake. While our model uses a context of 5, for simplicity, we show only 2 previous turns.
}
\label{table:intentExamples}
\end{table*}

We found great improvements for the `ask-attribute' intent where the KB-grounded model could answer correctly questions related to brand or colour and other attributes of the product, which resulted in an increase of 10 \textsc{Bleu} points on test instances with this user intent (M-HRED--attn--kb compared to M-HRED--attn). Similarly, in the example related to the `buy' intent in Table~\ref{table:intentExamples}, our model is able to learn that the product bought by the user is `kurta', which probably cannot be captured by the visual features. Hence, M-HRED--attn produces `jeans' on the output. M-HRED--attn--kb on the other hand learns this information from the KB. We also found that our \textsc{Bleu} score for the `show-orientation' intent has decreased w.r.t.\ to the non-KB-grounded model. A detailed probe found that the orientations for retrieved images may not directly follow the description in the query (KB). There are other intents for which even KB does not help, such as those requiring user modelling. 

\section{Conclusion and Future Work}
\label{sect:ConclusionFutureWork}
This work focuses on the task of textual response generation in multimodal task-oriented dialogue system. We used the recently released Multimodal Dialogue (MMD) dataset \cite{saha2017multimodal} for experiments and introduced a novel conversational model grounded in language, vision and Knowledge Base (KB). Our best performing model outperforms the baseline model \cite{saha2017multimodal} by 9 \textsc{Bleu} points, improving context modelling in multimodal dialogue generation. Even though our model outputs showed a substantial improvement (over 3 \textsc{Bleu} points) on incorporating KB information, integrating visual context still remains a bottleneck, as also observed by \citet{agrawal2016analyzing,qian2018multimodal}. This suggests the need for a better mechanism to encode visual context. 

Since our KB-grounded model assumes user intent annotation and KB queries as additional inputs, we plan to build a model to provide them automatically. 


\section*{Acknowledgments}
This research received funding from Adeptmind Inc., Toronto, Canada and the MaDrIgAL EPSRC project
(EP/N017536/1). The Titan Xp used for this
work was donated by the NVIDIA Corp.
%
\bibliography{biblio}
\bibliographystyle{acl_natbib}

\end{document}